\pdfoutput=1

    \PassOptionsToPackage{numbers, compress}{natbib}

\documentclass[11pt]{article}

\usepackage[preprint]{nips}

\usepackage{times}
\usepackage{latexsym}

\usepackage{booktabs}
\usepackage{multirow}
\usepackage{subfigure}
\usepackage{graphicx}
\usepackage{tabularx}
\usepackage{url}

\usepackage[T1]{fontenc}

\usepackage[utf8]{inputenc}

\usepackage{microtype}

\usepackage{inconsolata}

\usepackage{amssymb}
\usepackage{amsmath}
\usepackage{amsthm}

\usepackage[backref]{hyperref}

\title{VECO 2.0: Cross-lingual Language Model Pre-training with Multi-granularity Contrastive Learning}

\author{Zhen-Ru Zhang  \space\space Chuanqi Tan  \space\space Songfang Huang \space\space Fei Huang \\
    Alibaba DAMO Academy \\
  {\texttt{\{zhangzhenru.zzr,chuanqi.tcq,songfang.hsf,f.huang\}}@alibaba-inc.com}}

\begin{document}
\maketitle
\begin{abstract}

Recent studies have demonstrated the potential of cross-lingual transferability by training a unified Transformer encoder for multiple languages. In addition to involving the masked language model objective, existing cross-lingual pre-training works leverage sentence-level contrastive learning or plugs in extra cross-attention module to complement the insufficient capabilities of cross-lingual alignment. Nonetheless, synonym pairs residing in bilingual corpus are not exploited and aligned, which is more crucial than sentence interdependence establishment for token-level tasks. In this work, we propose a cross-lingual pre-trained model VECO~2.0 based on contrastive learning with multi-granularity alignments. Specifically, the sequence-to-sequence alignment is induced to maximize the similarity of the parallel pairs and minimize the non-parallel pairs. Then, token-to-token alignment is integrated to bridge the gap between synonymous tokens excavated via the thesaurus dictionary from the other unpaired tokens in a bilingual instance. Experiments show the effectiveness of the proposed strategy for cross-lingual model pre-training on the XTREME benchmark\footnote{Rank 1st on March 17, 2023 on XTREME leaderboard. \url{https://sites.research.google/xtreme/}}.

\end{abstract}

\section{Introduction}
Pre-trained models play an important role as a backbone for various NLP downstream tasks. The models have expanded from monolingual to multilingual with development, where cross-lingual pre-trained models have demonstrated their superior performance on cross-lingual NLP tasks \cite{devlin-etal-2019-bert, chi-etal-2022-xlm, conneau-etal-2020-unsupervised}.

To construct the universal representation between different languages, previous works mainly focus on two pre-training objectives, which are Multilingual Masked Language Model (MMLM) and Translation Language Model (TLM). MMLM is the multilingual version of MLM modeling each language separately in the shared semantic space and TLM performs MLM on concatenated parallel sentence pairs to implicitly capture the alignment via attention mechanism, both of them align the masked tokens with the context without considering sentence-level information. To overcome that, HICTL \cite{wei2021learning} and infoXLM \cite{chi-etal-2021-infoxlm} incorporate sentence-level contrastive learning to enhance the alignment among parallel sentences. However, one potential issue lies in that the token-to-token alignment for synonyms hidden in parallel corpus is ignored and lacks exploitation, despite sequence-to-sequence and token-to-sequence exploration have been included in the above approach, especially token alignment across languages is more crucial for token-oriented downstream tasks, i.e. cross-lingual Named Entity Recognition (NER). Furthermore, instead of implicitly building interdependence in TLM relying on self-attention module, VECO \cite{luo-etal-2021-veco} plugs a cross-attention module into Transformers encoder to explicitly capture alignment whose extra architecture has to be adapted and leads to extra parameters.

In light of the above motivation, we propose \textbf{V}arious granularity aligned \textbf{E}ncoder with \textbf{CO}ntrastive Learning (\textbf{VECO 2.0}) in sequence-to-sequence and token-to-token alignments. Specifically, VECO~2.0 maximizes the similarity of the parallel pairs and minimizes the non-parallel pairs in a batch based on the sequence semantic representation of bilingual corpus for sequence-to-sequence alignment. Besides, the synonyms residing in the parallel pairs are excavated via the thesaurus dictionary to construct the parallel token pairs. Similarly, VECO 2.0 bridges the gap between token pairs while separating from the other unpaired tokens in the instance. The above strategy is implemented on Transformer encoder architecture that can be directly adapted and combined with MLM and TLM tasks to establish the comprehensive alignment for token-sequence, sequence-sequence and token-token level, resulting in the universal representation cross languages.

We evaluate VECO 2.0 on a variety of representative cross-lingual NLU tasks in XTREME \cite{pmlr-v119-hu20b} benchmark including tasks of sentence-pair classification, structured prediction, question answering and sentence retrieval. Comparative experiments against multiple cross-lingual pre-trained models clearly demonstrate the effectiveness and superiority of our model. In addition, the ablation study further validates the two auxiliary alignment tasks play a crucial role in the sequence-level and token-level downstream tasks, respectively, guiding us in taking the task characteristics into account when pre-training. Finally, we pre-train a larger-scaled model based on the proposed mechanism, coupled with the fine-tuning and ensemble strategy, ranking first on the XTREME leaderboard on March 17, 2023. 

\begin{figure*}[]
    \centering
    \includegraphics[width=0.9 \textwidth]{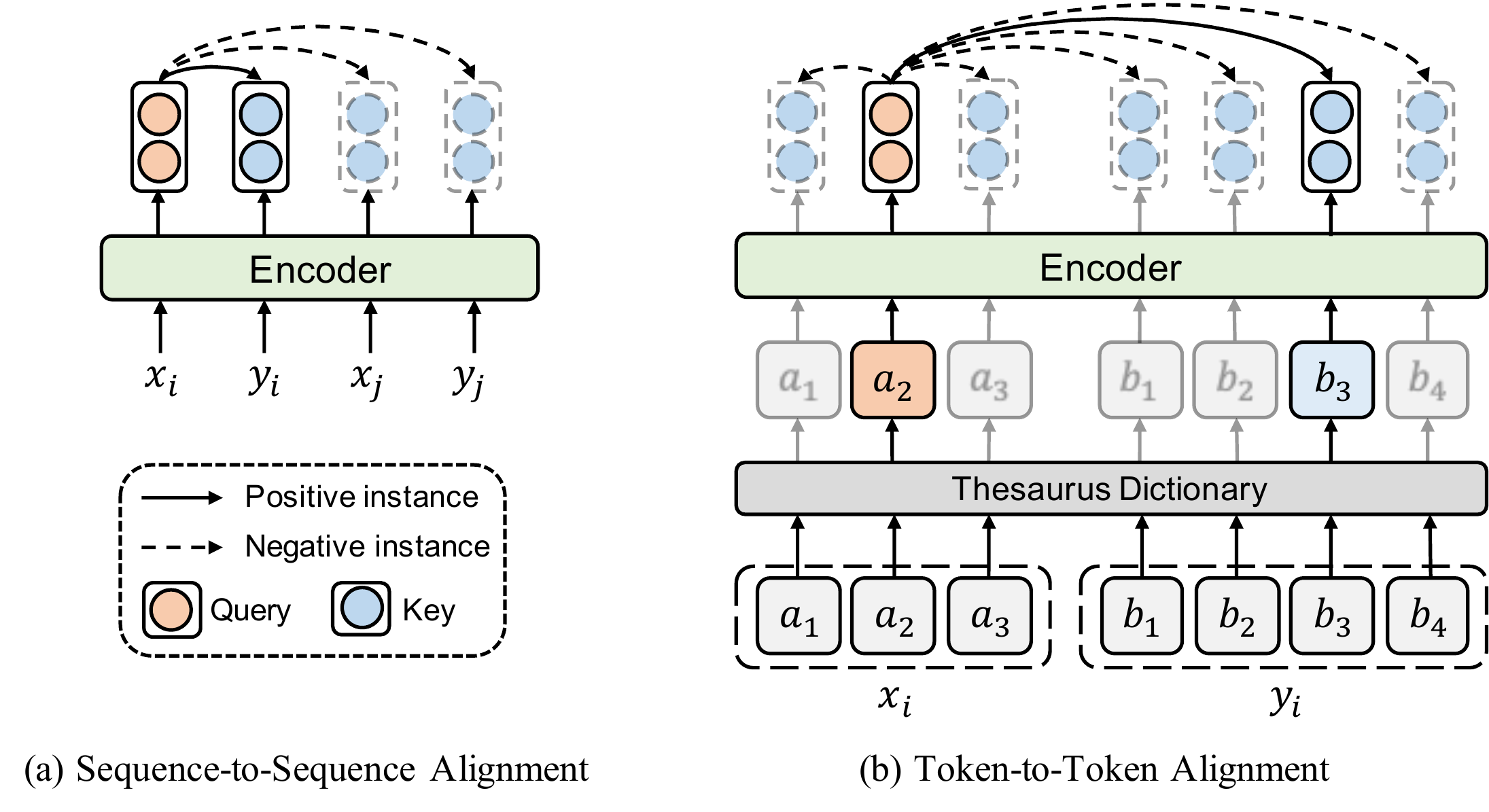}
    \caption{An illustration of the proposed multi-granularity contrastive learning where the left is the sequence-to-sequence alignment for a batch, and the right is the token-to-token alignment for an instance. In (a), $(x_i, y_i)$ indicates the parallel pair in a batch and $(x_j, y_j)$ is the another pair. In (b), $\{a_1,a_2,a_3\}$ and $\{b_1, b_2,b_3, b_4\}$ denote the tokens of $x_i$ and $y_i$, respectively. $(a_2, b_3)$ is a synonymous token pair filtered by thesaurus dictionary.}
    \label{pretrain model structure}
\end{figure*}

\section{Related Work}
\subsection{Multilingual Language Models}
Most existing multilingual language models are built based on Transformer encoder \cite{NIPS2017_3f5ee243} architecture. Among them, mBERT \cite{devlin-etal-2019-bert} is the first multilingual language model which constructs the sharing vocabulary and performs masked language modeling (MLM) task on monolingual corpus among multiple languages. XLM \cite{lample2019cross} introduces parallel data and extends MLM to translation language modeling (TLM) which randomly masks words in concatenated parallel sentences. XLM-R \cite{conneau-etal-2020-unsupervised} builds a larger vocabulary size of 250k and trains with multilingual MLM objective for only monolingual data in scaling amount based on RoBERTa \cite{liu2019roberta} architecture. InfoXLM \cite{chi-etal-2021-infoxlm} adds sentence-level contrastive learning loss for bilingual pairs to maximize the mutual information between translation pairs. ERNIE-M \cite{ouyang-etal-2021-ernie} integrates back-translation into the pre-training process to generate pseudo-parallel pairs for monolingual corpus, enabling alignment between different languages. HICTL \cite{wei2021learning} proposes sentence-level and word-level contrastive learning to distinguish the parallel sentence and related words for each sentence. XLM-E \cite{chi-etal-2022-xlm} introduces ELECTRA-style \cite{Clark2020ELECTRAPT} tasks including multilingual replaced token detection and translation replaced token detection. XY-LENT \cite{patra2022englishcentric} leverages X-Y bitexts coupled with a novel sampling strategy rather than previous English-centric bitexts.

Furthermore, there are also various multilingual language models built on Transformer encoder-decoder architecture which focus on the improvement of text generation and machine translation. For instance, mBART \cite{liu-etal-2020-multilingual-denoising} is a sequence-to-sequence denoising model pre-trained on monolingual corpora using the BART autoregressive objective. mT5 \cite{xue-etal-2021-mt5} introduces a multilingual variant of T5 \cite{2020t5} and significantly improves the performance. VECO \cite{luo-etal-2021-veco} provides a cross-attention module to build the interdependence between languages, which can be used to initialize both encoder-decoder models for NLG tasks and encoder models for NLU tasks.

Despite contrastive learning being utilized in our encoder model similar to infoXLM and HICTL, the key difference from previous work lies in that we construct the multi-granularities contrastive loss for alignment. Compared to the individual sentence-level contrastive learning in infoXLM, we also add token-level contrastive loss for synonym alignment. In contrast to word-level contrastive learning in HICTL which enhances the connection between related words and sentences for token-to-sequence alignment, VECO 2.0 bridges the representations of synonyms pairs embedded in the bilingual corpus rather than the tokens and the sentence they belong to, resulting in better performance in token-level cross-lingual downstream tasks.

\subsection{Contrastive Learning}
\textbf{C}on\textbf{t}rastive \textbf{L}earning (CTL) has been applied and validated in the field of computer vision \cite{he2020momentum,chen2020simple,chen2020improved,chen2020big}, then transferred to natural language processing \cite{gao-etal-2021-simcse,fang2020cert,wu2020clear,giorgi-etal-2021-declutr}. The key idea of that is drawing positive pairs closer while separating the negative pairs by a contrastive loss, where the construction of pairs and loss function definition is significant, enabling models to learn better representation. SimCLR \cite{chen2020simple} is the classic in-batch CTL method that builds the augmented images as positive pairs and the other unpaired images within the batch compose negative pairs. Meanwhile, the loss function infoNCE \cite{oord2019representation} is adapted as optimization objectives. In the NLP areas, simCSE \cite{gao-etal-2021-simcse} leverages the various representation via dropout encoded by language model to serve as the positive pairs. CERT \cite{fang2020cert} designs self-supervised CTL task to learn sentence-level representation, complementing the previous token-level task in pre-training, such as MLM in BERT \cite{devlin-etal-2019-bert}. The specific CTL method is augmenting input with back translation \cite{sennrich-etal-2016-improving} and following MoCo \cite{he2020momentum} which maintains a queue of negative candidate instances.

\section{Pre-training}

VECO 2.0 is the upgrade language model of VECO \cite{luo-etal-2021-veco} which enhances the parallel corpus alignment cross granularities. Different from VECO plugging an additional cross-attention module into the Transformer encoder to explicitly build the interdependence between languages, VECO 2.0 is under the encoder-only architecture with fewer parameters. Accordingly, VECO 2.0 proposes the \textbf{M}ulti-granularity \textbf{C}on\textbf{t}rastive \textbf{L}earning (MCTL), i.e., sequence-to-sequence and token-to-token alignment auxiliary tasks to tackle the representations across languages. Besides vanilla Masked Language Modeling (MLM) task for monolingual corpus and Translation Language Modeling (TLM) task for bilingual corpus, the two multi-granularity auxiliary tasks for parallel corpus are illustrated as Figure~\ref{pretrain model structure}.

\subsection{Sequence-to-Sequence Alignment}
An in-batch sequence-to-sequence contrastive loss is designed to bridge the gap of parallel corpus and widen the distance between unpaired sentences in semantic space from a coarse-grained perspective. Formally, let $\mathcal{X}=\{x_i\mid 1 \leq i \leq n \}$ and $\mathcal{Y}=\{y_i\mid 1 \leq i \leq n \}$ be the instances set in a training batch respectively, where $n$ is the batch size and $y_i$ corresponds to the translation of $x_i$. For pair $(x_i, y_i)$, taking the encoded representation of $x_i$ as query, and other instances except $x_i$ in a batch, i.e. $\mathcal{X}^{\setminus x_i} \cup \mathcal{Y}$ is considered as keys. Since $y_i$ is the positive samples corresponding to $x_i$, the contrastive loss for $x_i$ can be calculated as followed:
\begin{eqnarray}\label{sequence contrastive loss for x}
l_{ctl}(x_i) = -\log \frac{\exp(s(x_i, y_i) / \tau)}{\sum\limits_{k \in \mathcal{X}^{\setminus x_i}\cup\mathcal{Y}} \exp(s(x_i, k) / \tau)}
\end{eqnarray}

Here, $\tau$ is the temperature parameter and $s(x, y)$ reflects similarity between $x$ and $y$, where we initialize it with cosine similarity $s(x, y)=\frac{x\cdot y}{\|{x}\| \cdot \|{y}\|}$. Symmetrically and similarly, $y_i$ is also assumed as a query and the candidate keys are included in $\mathcal{X} \cup \mathcal{Y}^{\setminus y_i}$. On the condition that $x_i$ is the positive sample for $y_i$, the contrastive loss for $y_i$ is constructed with the following formula:
\begin{eqnarray}\label{sequence contrastive loss for y}
l_{ctl}(y_i) = -\log \frac{\exp(s(y_i, x_i) / \tau)}{\sum\limits_{k \in \mathcal{X}\cup\mathcal{Y}^{\setminus y_i}} 
\exp(s(y_i, k) / \tau)}
\end{eqnarray}
Accordingly, the sequence-to-sequence contrastive loss for a training batch in size $n$ is determined by the following:
\begin{eqnarray}\label{sequence contrastive loss}
\mathcal{L}_{seq} = \frac{1}{2n} \sum_{i=1}^{n} \{l_{ctl}(x_i) + l_{ctl}(y_i)\}
\end{eqnarray}

\subsection{Token-to-Token Alignment}
Although sequence-to-sequence alignment is proposed for better overall cross-linguistic representation, there are still many synonyms residing in the translated pairs waiting to be fully exploited, which act as anchor words in some downstream tasks (e.g. NER). Motivated by that, a strategy for token-to-token alignment is investigated in this paper. Specifically, for parallel corpus $(x_i, y_i)$, we assume that there are several synonym pairs that can be exploited via the mapping and filter from thesaurus dictionary as shown in Figure \ref{pretrain model structure}. Let $\mathcal{S}_i=\{(a_j, b_j)\mid a_j \in x_i, b_j \in y_i, 1 \leq j \leq |\mathcal{S}|\}$ be the set of synonym pairs, where $a_j$ and $b_j$ are positive samples with each other. The remaining tokens in the instance $x_i$ and $y_i$ are served as negative samples. Formally speaking, let $\mathcal{W}$ be the set of tokens from $x_i$ and $y_i$, and the contrastive loss for $a_j$ and $b_j$ are defined as followed separately.

\begin{eqnarray}\label{token contrastive loss for a}
l_{ctl}(a_j) = -\log \frac{\exp(s(a_j, b_j) / \tau)}{\sum\limits_{k \in \mathcal{W}^{\setminus a_j}} \exp(s(a_j, k) / \tau)}
\end{eqnarray}

\begin{eqnarray}\label{token contrastive loss for b}
l_{ctl}(b_j) = -\log \frac{\exp(s(b_j, a_j) / \tau)}{\sum\limits_{k \in \mathcal{W}^{\setminus b_j}} \exp(s(b_j, k) / \tau)}
\end{eqnarray}
Then, the token-to-token loss for the parallel pair $(x_i, y_i)$ can be derived via the aggregation of token pairs $\mathcal{S}_i$:
\begin{eqnarray}\label{token contrastive loss for xy pair}
l_{tok}(x_i, y_i) = \frac{1}{2|\mathcal{S}_i|} \sum_{j=1}^{|\mathcal{S}_i|} \{l_{ctl}(a_j)+l_{ctl}(b_j)\}
\end{eqnarray}
Finally, the token-to-token alignment loss for a training batch is determined by averaging loss for all bilingual pairs:
\begin{eqnarray}\label{token contrastive loss}
\mathcal{L}_{tok} = \frac{1}{n} \sum_{i=1}^{n} l_{tok}(x_i, y_i)
\end{eqnarray}

\subsection{Pre-training Tasks}
For monolingual corpus, VECO 2.0 employs MLM task which randomly replaces the token with \texttt{[MASK]} to align it with the context in its own language. For bilingual data, the TLM task which concatenates parallel sentences and performs MLM objective is utilized to implicitly attend to a part of words across languages. In addition, the above multi-granularity alignment tasks are also combined to capture cross-language correlation explicitly. Formally, the entire loss for the training corpus is optimized as followed:
\begin{eqnarray}\label{pretrain loss}
\mathcal{L}=\begin{cases}
\hspace{30pt}\mathcal{L}_{MLM}&, ~~{\rm if}~~{\rm monolingual} \\
\mathcal{L}_{TLM} + \mathcal{L}_{seq} + \mathcal{L}_{tok}&, ~~{\rm if}~~{\rm bilingual}
\end{cases}
\end{eqnarray}

\begin{table*}[]
\centering
\resizebox{1\textwidth}{!}{
\begin{tabular}{@{}lcccccll@{}}
\toprule
Model & Architecture & \#Params. & \#Layers & \#Langs. & \#Vocab. & Tasks & Training Data \\ \midrule
mBERT & Encoder & 172M & 12 & 104 & 110k & MLM & Wikipedia \\
XLM & Encoder & 570M & 24 & 100 & 200k & MLM,TLM,CTM & Wikipedia+Translation \\
XLM-R & Encoder & 559M & 24 & 100 & 250k & MLM & CommonCrawl \\
HICTL & Encoder & 559M & 24 & 100 & 250k & MLM, TLM, HICTL & CommonCrawl+Translation \\
VECO & Flexible & 662M & 24$^{*}$ & 50 & 250k & MLM, TLM, CA-MLM & CommonCrawl+Translation \\
VECO 2.0 & Encoder & 559M & 24 & 109 & 250k & MLM, TLM, MCTL & CommonCrawl+Translation \\ \bottomrule
\end{tabular}
}
\caption{The details of the compared cross-lingual model. * denotes VECO unifies the encoder and decoder.}
\label{model details}
\end{table*}


\section{Experiments}

\subsection{Pre-training Corpus}
We pre-train on monolingual and bilingual corpus involving 109 languages. For monolingual data, we follow XLM-R \cite{conneau-etal-2020-unsupervised} using CC-100\footnote{https://data.statmt.org/cc-100/} \cite{wenzek-etal-2020-ccnet} from Common Crawl and extract 2.5TB data. For bilingual data, we collect 4TB parallel pairs from OPUS\footnote{https://opus.nlpl.eu/}.

To mitigate the imbalance between high and low-resource languages, we sample monolingual corpus with the multinomial distribution following XLM \cite{conneau2019cross}. Specifically, in $N$ languages, the sampling probability $q_{i}$ for language $i~(1<j<N)$ can be formalized as follow:
\begin{eqnarray}\label{language sample}
q_{i} = \frac{p_{i}^{\alpha}}{\sum_{j=1}^N {p_{j}^{\alpha}}},~{\rm where}~~p_{i} = \frac{n_{i}}{\sum_{k=1}^N {n_{k}}}
\end{eqnarray}
Here, $n_{i}$ is the number of sentences for language $i$ and $\alpha$ corresponds to the smoothing parameter which controls the language sampling rate. The lower the value of $\alpha$, the more inclined the low resource language. We employ $\alpha=0.5$.

\subsection{Implementation Details}
The model is large-scaled that has 559M parameters in 24 layers with 1024 hidden size and 4096 feed-forward size. We adopt the 250k shared vocabulary same as XLM-R and apply subword tokenization directly on raw text data using Sentence Piece Model \cite{kudo-richardson-2018-sentencepiece}. Following XLM-R\cite{conneau-etal-2020-unsupervised} and VECO \cite{luo-etal-2021-veco}, we do not use language embedding for better generalization. Table \ref{model details} shows the details of the involved model. The model parameters are initialized by the encoder of VECO. During the training phase, to balance the monolingual and bilingual corpus, we alternately sample a batch of monolingual segments and a batch of parallel sentences. In the token-to-token alignment task, The MUSE\footnote{https://github.com/facebookresearch/MUSE} \cite{conneau2018word} is utilized as the thesaurus dictionary.

\subsection{Downstream Tasks}
In this paper, we evaluate our model on XTREME \cite{pmlr-v119-hu20b} which is a massively multilingual benchmark for evaluating cross-lingual generalization. Specifically, XTREME includes 9 tasks from 4 categories covering 40 languages:
\begin{itemize}
\item Sentence-pair classification: Cross-lingual Natural Language Inference (XNLI) \cite{conneau-etal-2018-xnli} and Cross-lingual Paraphrase Adversaries from Word Scrambling (PAWS-X) \cite{zhang-etal-2019-paws}. XNLI aims to predict the relation between the premise and hypothesis sentence, i.e. entailment, contradiction or neutral. PAWS-X determines whether the two sentences are paraphrased of each other.
\item Structured prediction: POS tagging from the Universal Dependencies v2.5 \cite{nivre-etal-2020-universal} treebanks which each word is assigned one of 17 universal POS tags and NER from Wikiann \cite{pan-etal-2017-cross} annotated with LOC, PER, and ORG tags in IOB2 format.
\item Question answering: Cross-lingual Question Answering (XQuAD) \cite{artetxe-etal-2020-cross}, Multilingual Question Answering (MLQA) \cite{lewis-etal-2020-mlqa}, and the gold passage version of the Typologically Diverse Question Answering dataset (TyDiQA-GoldP) \cite{clark-etal-2020-tydi}. They are all extractive QA where the answer spans concealed in the context.
\item Sentence retrieval: Building and Using Parallel Corpora (BUCC) \cite{zweigenbaum-etal-2017-overview} and Tatoeba dataset~\cite{artetxe-schwenk-2019-massively} which aim to extract parallel sentences between the English corpus and target languages.
\end{itemize}

Here, for cross-lingual setting, all tasks provide English training data and dev/test set in all involved languages, except sentence retrieval tasks which have no training data and are in a zero-shot setting.

\begin{table*}[tp]
\centering
\resizebox{1\textwidth}{!}{
\begin{tabular}{@{}lccccccccccccc@{}}
\toprule
 & \multicolumn{2}{c}{Pair sentence} & \multicolumn{2}{c}{Structured prediction} & \multicolumn{6}{c}{Question answering} & \multicolumn{2}{c}{Sentence retrieval} &  \\
\multirow{-2}{*}{Datasets} & XNLI & PAWS-X & POS & NER & \multicolumn{2}{c}{XQuAD} & \multicolumn{2}{c}{MLQA} & \multicolumn{2}{c}{TyDiQA} & BUCC & Tatoeba &  \\ \cmidrule(r){1-13}
\#Langs & 15 & 7 & 33 & 40 & \multicolumn{2}{c}{11} & \multicolumn{2}{c}{7}& \multicolumn{2}{c}{9} & 5 & 33 &  \\ \cmidrule(r){1-13}
Metrics & Acc. & Acc. & F1 & F1 & F1 & EM & F1 & EM & F1 & EM & F1 & Acc. & \multirow{-4}{*}{AVG} \\ \midrule
\multicolumn{14}{l}{\textit{Cross-lingual zero-shot transfer (models are trained on English data)}} \\ \midrule
mBERT & 65.4 & 81.9 & 71.5 & 62.2 & 64.5 & 49.4 & 61.4 & 44.2 & 59.7 & 43.9 & 56.7 & 38.7 & 59.8 \\
XLM & 69.1 & 80.9 & 71.3 & 61.2 & 59.8 & 44.3 & 48.5 & 32.6 & 43.6 & 29.1 & 56.8 & 32.6 & 55.7 \\
XLM-R & 79.2 & 86.4 & 73.8 & 65.4 & 76.6 & 60.8 & 71.6 & 53.2 & 65.1 & 45.0 & 66.0 & 57.3 & 68.2 \\
HICTL & \textbf{81.0} & 87.5 & 74.8 & \underline{66.2} & \underline{77.9} & 61.7 & \textbf{72.8} & \textbf{54.5} & 66.0 & 45.7 & 68.4 & 59.7 & 69.6 \\
VECO & 79.9 & \textbf{88.7} & \underline{75.1} & 65.7 & 77.3 & \underline{61.8} & 71.7 & 53.2 & \underline{67.6} & \underline{49.1} & \underline{85.0} & \underline{75.1} & \underline{73.1} \\
VECO 2.0 & \underline{80.4} & \underline{88.5} & \textbf{75.4} & \textbf{67.2} & \textbf{78.9} & \textbf{63.7} & \underline{72.7} & \underline{54.3} & \textbf{71.1} & \textbf{54.7} & \textbf{86.2} & \textbf{81.8} & \textbf{75.2} \\
\midrule
\multicolumn{14}{l}{\textit{XTREME leaderboard}} \\ \midrule
VECO 2.0 & 88.3 & 93.4 & 85.3 & 84.0 & 85.9 & 73.2 & 80.5 & 63.9 & 85.4 & 74.2 & 93.8 & 96.2 & 85.8 \\ \bottomrule
\end{tabular}
}
\caption{XTREME results on each dataset. The results of mBERT, XLM and XLM-R are from \cite{pmlr-v119-hu20b}, and those for VECO and HICTL are from their respective papers \cite{luo-etal-2021-veco} and \cite{wei2021learning}. The detailed results for each language are in Appendix \ref{sec:appendix of detailed results on XTREME}. (\textbf{bold}: the best score; \underline{underline}: the second.)}
\label{xtreme res}
\end{table*}

\subsection{Fine-tuning Setting}
We consider the most common cross-lingual setting for evaluation of the cross-lingual language model. Specifically, only the English training corpus is used for fine-tuning and directly evaluating on dev/test dataset in other languages, aiming to measure the cross-lingual transfer capability of the model. We adopt the fine-tuning training scripts provided by XTREME \footnote{https://github.com/google-research/xtreme}. The hyperparameters setting for fine-tuning are shown in Appendix \ref{sec:appendix of hyperparameters for fine-tuning}.

\subsection{Results}
As shown in Table \ref{xtreme res}, we evaluate VECO 2.0 on the nine tasks of XTREME in the cross-lingual setting compared with mBERT \cite{devlin-etal-2019-bert}, XLM \cite{chi-etal-2022-xlm}, XLM-R \cite{conneau-etal-2020-unsupervised}, HICTL \cite{wei2021learning} and VECO \cite{luo-etal-2021-veco}. It can be found that VECO 2.0 outperforms the mBERT, XLM, XLM-R and VECO on all nine tasks of 12 metrics with an average improvement of +15.4\%, 19.5\%, +7.0\%, 2.1\%, surpasses HICTL on 9/12 metrics with an average advancement of +5.6\%, respectively.

Specifically, for sentence-pair classification tasks, the performance of VECO 2.0 on XNLI and PAWS-X is slightly inferior to HICTL and VECO by 0.6\%, 0.2\% separately, but still superior to other models, which we suppose the token-to-token alignment affects the sentence-to-sentence alignment to some extent in the condition that XNLI focuses on the sentence-level semantic representation. We further validate that in the ablation study. For structured prediction, VECO 2.0 exceeds the sub-optimal performance by 0.3\% with VECO on POS tagging and 1.0\% with HICTL on NER. The advantageous performance of NER we attribute to the potential of the token-to-token alignment task to bridge the distance between synonymous cross-lingual entities in the semantic space. For question answering, VECO 2.0 achieves the best performance, surpassing VECO by 1.75\% and HICTL by 1.5 \% in the average of F1 and EM on XQuAD. For the average of F1 and EM on MLQA, VECO 2.0 outperforms VECO by 1.05\% and is slightly inferior to HICTL by 0.15\%. On TyDiQA-GoldP, VECO 2.0 achieves significant gains by a wide margin against other models, where over VECO 4.55\% and HICTL 7.05\%. For sentence retrieval, both VECO and VECO 2.0 significantly improve performance on two retrieval tasks compared to other models, but VECO 2.0 further raises it by over VECO 1.2\% on BUCC and 6.7\% on Tatoeba respectively. In summary, our method delivers the best overall performance. We conclude the reasons for improvement lie in the strong representation alignment and interdependence establishment across different languages via the proposed multi-granularity contrastive learning tasks, which will be further investigated in the ablation study.

\begin{table*}[]
\center
\small
\begin{tabular}{@{}lcccccccc@{}}
\toprule
Model & XNLI & PAWS-X & XQuAD & MLQA & TyDiQA & NER & POS & Avg. \\ \midrule
mBERT & 16.5 & 14.1 & 25 & 27.5 & 22.2 & 23.6 & 25.5 & 22.1 \\
XLM-R & 10.2 & 12.4 & 16.3 & 19.1 & 13.3 & 19.8 & 24.3 & 16.5 \\
XLM & 14.7 & 13.1 & 19.6 & 26.3 & 27.9 & 22.0 & 24.8 & 21.2 \\
VECO & \textbf{8.9} & 7.5 & 16.6 & 20.2 & 10.2 & 18.5 & \textbf{21.4} & 14.8 \\
VECO 2.0 & 9.2 & \textbf{7.3} & \textbf{16.2} & \textbf{20.1} & \textbf{6.8} & \textbf{18.1} & \textbf{21.4} & \textbf{14.1} \\ \bottomrule
\end{tabular}
\caption{The cross-lingual transfer gap of different pre-trained models on XTREME tasks. The transfer gap is the difference between performance on the English test set and the average performance in the other languages, where the lower score the better transferability. For the QA tasks, we show EM scores. (\textbf{bold}: the best scores.)}
\label{cross-lingual transfer gap exp}
\end{table*}

\begin{table*}[b]
\center
\resizebox{1\textwidth}{!}{
\begin{tabular}{@{}lccccccc@{}}
\toprule
Tasks & XNLI & POS & NER & \multicolumn{2}{c}{TyDiQA-GoldP} & BUCC & \multirow{2}{*}{AVG} \\
Metrics & Acc. & F1 & F1 & EM & F1 & F1 &  \\ \midrule
MLM+TLM & 71.1 & 67.0 & 53.5 & 32.6 & 48.8 & 23.4 & 49.4 \\
\midrule
MLM+TLM+Seq-Seq CTL & 72.3$_{(+1.2)}$ & 67.4$_{(+0.3)}$ & 53.9$_{(+0.4)}$ & 33.0$_{(+0.4)}$ & 49.8$_{(+1.0)}$ & 48.9$_{(+25.5)}$ & 54.2$_{(+4.8)}$ \\
MLM+TLM+Tok-Tok CTL & 70.2$_{(-0.9)}$ & 67.1$_{(+0.1)}$ & 54.4$_{(+0.9)}$ & 31.9$_{(-0.7)}$ & 49.3$_{(+0.5)}$ & 24.4$_{(+0.9)}$ & 49.5$_{(+0.1)}$ \\
MLM+TLM+MCTL & 71.3$_{(+0.2)}$ & 68.1$_{(+1.1)}$ & 55.5$_{(+2.0)}$ & 33.7$_{(+1.1)}$ & 51.4$_{(+2.6)}$ & 34.9$_{(+11.4)}$ & 52.5$_{(+3.1)}$ \\
\bottomrule
\end{tabular}
}
\caption{Ablation study of base-sized VECO 2.0. We gradually add alignment tasks on the basis of MLM + TLM with consistent training data and hyperparameters. MCTL indicates that both Seq-Seq CTL and Tok-Tok CTL are used. The number in $(\cdot)$ reflects the difference between the current setting and the baseline MLM + TLM.}
\label{ablation study}
\end{table*}

\subsection{Analysis}

\subsubsection{Cross-lingual Transfer Gap}
The cross-lingual transfer gap \cite{pmlr-v119-hu20b} is the difference between performance on the English test set and the average performance on the other languages. The lower transfer gap indicates the better cross-lingual transfer capability of the model and a transfer gap of 0 suggests the perfect cross-lingual transfer. Table \ref{cross-lingual transfer gap exp} demonstrates the comparison of VECO 2.0 and the other cross-lingual pre-trained model on XTREME tasks, which can be summarized that VECO 2.0 have the better cross-lingual transferability on average among them. In particular, combined with Table \ref{xtreme res}, VECO 2.0 has not only better average performance across all languages on XQuAD, MLQA, TyDiQA and NER but also a lower transfer gap on them against VECO. For task XNLI, our performance is better than VECO, but the transfer gap is higher, which we suggest probably is caused by overfitting the English data. While the condition is reversed for task PAWS-X, in other words, VECO 2.0 has greater transferability although the performance is slightly inferior to VECO.

\subsubsection{Ablation Study}
To explore the impact of each alignment loss, we conducted a series of ablation experiments as presented in Table \ref{ablation study}. We pre-train the same steps on the base scaled model across various settings using monolingual and bilingual corpus in selected languages, with the only difference being the training objective. Specifically, we compared four settings, starting with the vanilla MLM for monolingual data and TLM for parallel data without any auxiliary alignment tasks, and subsequently incorporating sequence-to-sequence and token-to-token CTL loss in each order. To sum up, the following observations can be made based on the reported results.

First, sequence alignment has a significant impact on improving the retrieval task BUCC, next is the classification task XNLI, which suggests that sequence alignment can effectively bridge the semantic representation of parallel data. Second, the token alignment task demonstrates better improvement for token-level downstream tasks, i.e. NER, than the sentence alignment task. But it hurts the performance of sentence-level tasks like XNLI to some extent, which we attribute to the potential disruption of overall sentence representation caused by the attraction between synonyms. Third, when sequence and token alignment are used in conjunction denoted as MCTL, we observe a general boost compared to vanilla MLM+TLM, which is more pronounced and evenly distributed than using only sequence alignment task without considering the effect of extreme values brought from BUCC. Besides, for the QA task TyDiQA-Gold, it can be observed that combining two auxiliary tasks jointly works better than separate training, where we attribute this improvement to the fact that the QA task requires both an overall sentence semantic understanding of the question and the segmentation of the answer token corresponding to the sequence and token alignment respectively.

In general, sentence-level downstream tasks, i.e. classification and retrieval require sentence alignment whereas token alignment is more crucial for token-level tasks like NER. For QA, the involvement of both sentence and token alignment is significant for achieving optimal results. Considering the variety and different characteristics of the downstream tasks in XTREME, we ultimately incorporate the two alignment tasks to achieve a greater and more consistent improvement.

\section{XTREME Leaderboard Submission}
Based on the key strategy of VECO 2.0, we submit our results coupled with effective fine-tuning and ensemble methods to the XTREME leaderboard. The details involved are shown below.
\subsection{Model Setting}
In addition to the large model, we also pre-train an xlarge scaled model of 3.5B parameters which has 36 layers with 2,560 hidden size and 10,240 feed-forward size. The pre-training corpus and shared vocabulary are kept consistent with the setting of large model. We initialize the parameters of the model with XLM-R xlarge \cite{goyal-etal-2021-larger}.

\subsection{Task Setting}

\paragraph{Translate-Train}
In the translate-train setting, the cross-lingual or multilingual pre-trained model is fine-tuned on the collection of all data, i.e. golden training corpus in English and the corresponding translated corpus in other languages. XTREME offers translated training corpus in other languages for translate-train setting and translated test corpus in English for translate-test setting for most tasks \footnote{https://github.com/google-research/xtreme}. Note that for two structure prediction tasks (POS, NER), the position of token labels in the translated text generally differs from that in the source text. For the POS task, we follow VECO \cite{luo-etal-2021-veco} that uses the model trained only on the
English training dataset as a teacher, to label the
translated text. For the NER task, we follow EASYPROJECT \cite{Chen2022FrustratinglyEL} that utilizes a simple mark-then-translate method to label the translated text by inserting special markers around the labeled spans in the original sentence. In practice, we additionally filter the translated text whose projected entities are equal to the prediction of models in the cross-lingual setting. It is found that the label of translate-train data filtered is of high quality, leading to the best experimental results.

\paragraph{Translate-Test}
In the translate-test setting, the pre-trained model is trained on the English training data and evaluated on test data translated from the
target language to English. Referring to the settings and results of XTREME \cite{pmlr-v119-hu20b}, we conduct experiments under translate-test settings in tasks of sentence-pair classification and question answering. Similar to the translate-train setting, XTREME also offers translated test corpus for these tasks. For sentence-pair classification, we predict the original and translated test data at the same time, then we consider both logits to obtain the final prediction via a certrain strategy, e.g. maximum or mean. For question answering, we first use a model trained on the English training dataset to predict the answer on the translate-test text, and then translate to answer to the target language by Google Translate. Note that there exists a mismatch between the translated answer and the span in original text. Unlike mapping the answer to the closest span, we find that using it as an anchor for selecting the top-k results can achieve better results. Therefore, in practice, we fine-tuned the pre-trained model under the translate-train setting to predict the top-k results on original test data, then we select the final answer by the token level similarity with the answer of the translate-test.

\subsection{Fine-tuning Strategy}

We leverage different fine-tuning strategies including Child-Tuning \cite{xu-etal-2021-raise}, Hype \cite{Yuan2022HyPeBP}, R-Drop \cite{liang2021rdrop}, and implement an alignment-enhanced fine-tuning method for parallel training corpus in translate-train setting, where the alignment loss between bilingual data is attached to the original downstream tasks loss. The specific loss function we use is either Kullback–Leibler divergence following R-drop, or the infoNCE loss in contrastive learning similar to the one we used in the pre-training phase. It is worth noting that sentence retrieval is a zero-shot task without fine-fining, but we found the performances of the model fine-tuned by related downstream tasks is better than direct inference on the pre-trained model, which is also validated in \cite{phang-etal-2020-english}. In practice, we leverage XNLI to assist with sentence retrieval tasks.

\subsection{Ensemble Setting}
We fine-tune with different hyper-parameters e.g. learning rates and random seeds to generate the ensemble results. Detailed settings and hyper-parameters are shown in Appendix. Due to the different characteristics of the downstream tasks, we used separate ensemble settings.

\noindent \textbf{Sentence-pair Classification} We consider all the candidate probabilities to decide the final answer including the prediction of translate-test data.

\noindent \textbf{Structured Prediction} We take the probabilities to decide the label at the token level. For the NER task, we additionally filter the illegal entity under the BIO scheme.

\noindent \textbf{Question Answering} We first ensemble the result at the span level under the translate-train setting, and then take the translate-test result into consideration to decide the final answer as mentioned above.

\noindent \textbf{Sentence Retrieval} We obtain the representation of each layer of the models, and ensemble the result at pair level to decide the final output.

Finally, we achieve the average results of 85.8 on 9 tasks as shown in Table \ref{xtreme res}, which ranked 1st on the XTREME leaderboard on March 17, 2023.

\section{Conclusion}
In this paper, we investigate a multi-granularity alignment task via contrastive learning for cross-lingual language model pre-training, where the synonymous sequence and tokens in the parallel corpus are exploited to bridge the gap of semantic representation between languages, establishing the comprehensive alignment for token-sequence, sequence-sequence and token-token level combined with MLM and TLM. Extensive experiments on the broad downstream tasks of XTREME show the effectiveness of the proposed model. At the same time, we reveal the strategies involved in the first rank of the XTREME leaderboard, hoping to inspire future work.

\section*{Acknowledgements}

We would like to express our sincere appreciation to Fuli Luo, Xiangpeng Wei, Wei Wang, Haiyang Xu for their valuable suggestions and contributions to this work.

\bibliography{anthology,custom}
\bibliographystyle{plain}

\appendix

\section{Pre-Training Corpus Details}
\label{sec:appendix of pre-training corpus details }
For bilingual data, we collect 4TB parallel pairs (English-to-X) from the OPUS website \footnote{http://opus.nlpl.eu/} covering 109 languages, including MIZAN, tico-19, IITB, UNPC, MultiUN, ECB, WikiMatrix, Europarl, OpenSubtitles, News-Commentary, EUconst, Bianet, EuroPat, Books, ELITR-ECA, infopankki, CCMatrix, Tatoeba-Challenge, Wikipedia, ParaCrawl, CCAligned, news-commentary-v11, DGT, TED2020, UN, EU-bookshop, GoURMET and Tilde. In the ablation study, we pre-trained the base-sized model with a subset of the monolingual corpus in language en, de, es, fr, ru, tr, th, vi, zh and bilingual pairs in language en-es, en-fr, en-ru.

\section{Hyperparameters for Pre-training}
As shown in Table \ref{pre-training hyperparameters}, we present the hyperparameters for pre-training VECO 2.0 large model.

\section{Hyperparameters for Fine-tuning}
\label{sec:appendix of hyperparameters for fine-tuning}
For fine-tuning VECO 2.0 on the XTREME benchmark, we grid search the hyperparameters for the cross-lingual transfer setting. The detailed settings are listed in Table \ref{fine-tuning hyperparameters}.

\section{Detailed Results on XTREME}
\label{sec:appendix of detailed results on XTREME}
The detailed results in different languages of each XTREME task in the cross-lingual transfer setting on all languages are listed in the following tables.

\begin{table*}[]
\centering
\begin{tabular}{@{}ll@{}}
\toprule
Pre-training Hyperparameters & Large \\ \midrule
Number of layers & 24 \\
Hidden Size & 1024 \\
FFN inner hidden size & 4096 \\
Attention heads & 16 \\
Attention head size & 64 \\
Embedding Size & 1024 \\
Mask percent (monolingual/ bilingual) & 15\% / 25\% \\
Learning Rate Schedule & Linear \\
Warmup steps & 25K \\
Learning Rate & 1e-4 \\
Adam $\epsilon$ & 1e-6 \\
Adam $\beta_1$ & 0.9 \\
Adam $\beta_2$ & 0.999 \\
Attention Dropout & 0.1 \\
Dropout & 0.1 \\
Weight Decay & 0.01 \\
Max Sequence Length (monolingual/bilingual) & 512/128 \\
Batch Size (monolingual/bilingual) & 2048/2048 \\
Train Steps & 500K \\
Total Parameters & 559M \\ \bottomrule
\end{tabular}
\caption{The pre-training hyperparameters.}
\label{pre-training hyperparameters}
\end{table*}

\begin{table*}[]
\resizebox{1\textwidth}{!}{
\begin{tabular}{@{}lcccccc@{}}
\toprule
 & XNLI & PAWS-X & POS & NER & XQuAD/MLQA & TyDiQA \\ \midrule
Batch size & 32 & 32 & \{8,16,32\} & \{8,32\} & \{16, 32\} & \{16,32\} \\
Learning rate & \{2,$\cdots$,8\}e-6 & \{8,9,10,20\}e-6 & \{6,7,9\}e-6 & \{6,7,9\}e-6 & \{5,8,10,20,30\}e-6 & \{5,8,10,20,30\}e-6 \\
Epochs & \{5, 10\} & \{5, 10\} & \{5, 10\} & \{5, 10\} & \{2,3,4\} & \{10,20,40\} \\ \bottomrule
\end{tabular}
}
\caption{The fine-tuning hyperparameters.}
\label{fine-tuning hyperparameters}
\end{table*}

\begin{table*}[]
\centering
\resizebox{1\textwidth}{!}{
\begin{tabular}{@{}lcccccccccccccccc@{}}
\toprule
Model & en & ar & bg & de & el & es & fr & hi & ru & sw & th & tr & ur & vi & zh & Avg. \\ \midrule
XLM-R & 88.7 & 77.2 & 83.0 & 82.5 & 80.8 & 83.7 & 82.2 & 75.6 & 79.1 & 71.2 & 77.4 & 78.0 & 71.7 & 79.3 & 78.2 & 79.2 \\
VECO & 88.2 & 79.2 & 83.1 & 82.9 & 81.2 & 84.2 & 82.8 & 76.2 & 80.3 & 74.3 & 77.0 & 78.4 & 71.3 & 80.4 & 79.1 & 79.9 \\
VECO 2.0 & 88.9 & 79.1 & 83.4 & 83.0 & 82.7 & 84.9 & 83.2 & 76.7 & 80.7 & 71.9 & 77.8 & 79.5 & 72.8 & 80.6 & 79.9 & 80.4 \\ \bottomrule
\end{tabular}
}
\caption{XNLI accuracy scores for each language in cross-lingual setting.}
\end{table*}

\begin{table*}[]
\small
\centering
\begin{tabular}{@{}lcccccccc@{}}
\toprule
Model & en & de & es & fr & ja & ko & zh & Avg. \\ \midrule
XLM-R & 94.7 & 89.7 & 90.1 & 90.4 & 78.7 & 79.0 & 82.3 & 86.4 \\
VECO & 96.2 & 91.3 & 91.4 & 92.0 & 81.8 & 82.9 & 85.1 & 88.7 \\
VECO 2.0 & 95.8 & 91.0 & 91.6 & 92.0 & 82.8 & 81.6 & 84.5 & 88.5 \\ \bottomrule
\end{tabular}
\caption{PAWS-X accuracy scores for each language in cross-lingual setting.}
\end{table*}

\begin{table*}[]
\centering
\resizebox{1\textwidth}{!}{
\begin{tabular}{@{}lccccccccccccccccc@{}}
\toprule
Model & af & ar & bg & de & el & en & es & et & eu & fa & fi & fr & he & hi & hu & id & it \\ \midrule
XLM-R & 89.8 & 67.5 & 88.1 & 88.5 & 86.3 & 96.1 & 88.3 & 86.5 & 72.5 & 70.6 & 85.8 & 87.2 & 68.3 & 76.4 & 82.6 & 72.4 & 89.4 \\
VECO & 88.3 & 67.4 & 87.4 & 88.5 & 86.7 & 95.9 & 89.0 & 87.8 & 75.1 & 70.9 & 86.2 & 88.9 & 67.5 & 76.2 & 82.9 & 72.9 & 89.9 \\
VECO 2.0 & 89.4 & 70.0 & 88.6 & 89.7 & 86.6 & 96.2 & 89.0 & 87.2 & 75.1 & 71.4 & 86.1 & 87.7 & 70.2 & 74.7 & 84.2 & 72.8 & 89.8 \\
\midrule
 & ja & kk & ko & mr & nl & pt & ru & ta & te & th & tl & tr & ur & vi & yo & zh & Avg. \\
\midrule
XLM-R & 15.9 & 78.1 & 53.9 & 80.8 & 89.5 & 87.6 & 89.5 & 65.2 & 86.6 & 47.2 & 92.2 & 76.3 & 70.3 & 56.8 & 24.6 & 25.7 & 73.8 \\
VECO & 31.4 & 79.3 & 53.1 & 84.3 & 89.8 & 88.3 & 90.2 & 64.3 & 85.8 & 48.0 & 93.7 & 77.2 & 69.2 & 58.1 & 26.2 & 39.4 & 75.1 \\
VECO 2.0 & 36.2 & 78.3 & 53.4 & 84.7 & 89.8 & 88.8 & 89.8 & 64.9 & 84.5 & 50.9 & 93.3 & 76.8 & 67.0 & 58.8 & 23.2 & 40.7 & 75.4 \\ \bottomrule
\end{tabular}
}
\caption{POS F1 scores for each language in cross-lingual setting.}
\end{table*}

\begin{table*}[]
\centering
\resizebox{1\textwidth}{!}{
\begin{tabular}{@{}lccccccccccccccccccccc@{}}
\toprule
Model & en & af & ar & bg & bn & de & el & es & et & eu & fa & fi & fr & he & hi & hu & id & it & ja & jv &  \\ \midrule
XLM-R & 84.7 & 78.9 & 53.0 & 81.4 & 78.8 & 78.8 & 79.5 & 79.6 & 79.1 & 60.9 & 61.9 & 79.2 & 80.5 & 56.8 & 73.0 & 79.8 & 53.0 & 81.3 & 23.2 & 62.5 &  \\
VECO & 83.8 & 77.5 & 48.2 & 83.9 & 77.2 & 79.4 & 79.3 & 75.4 & 80.4 & 68.3 & 68.2 & 80.6 & 80.1 & 55.0 & 71.0 & 80.9 & 52.9 & 81.7 & 19.4 & 63.2 &  \\
VECO 2.0 & 84.8 & 78.5 & 51.8 & 81.3 & 79.2 & 80.7 & 80.7 & 75.8 & 82.4 & 69.6 & 66.3 & 81.1 & 80.7 & 56.2 & 73.1 & 82.9 & 54.1 & 82.1 & 18.8 & 66.1 &  \\
\midrule
 & ka & kk & ko & ml & mr & ms & my & nl & pt & ru & sw & ta & te & th & tl & tr & ur & vi & yo & zh & Avg. \\
 \midrule
XLM-R & 71.6 & 56.2 & 60.0 & 67.8 & 68.1 & 57.1 & 54.3 & 84.0 & 81.9 & 69.1 & 70.5 & 59.5 & 55.8 & 1.3 & 73.2 & 76.1 & 56.4 & 79.4 & 33.6 & 33.1 & 65.4 \\
VECO & 67.1 & 51.2 & 59.9 & 63.4 & 65.0 & 70.0 & 56.1 & 83.4 & 83.1 & 71.3 & 70.5 & 60.5 & 56.2 & 1.4 & 71.3 & 80.4 & 69.3 & 76.0 & 37.4 & 29.1 & 65.7 \\
VECO 2.0 & 72.8 & 53.7 & 59.3 & 67.9 & 66.0 & 68.8 & 56.0 & 85.0 & 82.2 & 72.5 & 69.3 & 61.4 & 54.3 & 1.8 & 75.8 & 82.5 & 76.0 & 78.0 & 49.3 & 28.0 & 67.2 \\ \bottomrule
\end{tabular}
}
\caption{NER F1 scores for each language in cross-lingual setting.}
\end{table*}

\begin{table*}[]
\centering
\resizebox{1\textwidth}{!}{
\begin{tabular}{@{}lcccccccccccc@{}}
\toprule
Model & en & ar & de & el & es & hi & ru & th & tr & vi & zh & Avg. \\ \midrule
XLM-R & 86.5/75.7 & 68.6/49.0 & 80.4/63.4 & 79.8/61.7 & 82.0/63.9 & 76.7/59.7 & 80.1/64.3 & 74.2/62.8 & 75.9/59.3 & 79.1/59.0 & 59.3/50.0 & 76.6/60.8 \\
VECO & 87.6/76.5 & 73.6/56.1 & 79.8/62.2 & 79.6/61.6 & 81.2/61.6 & 74.7/57.6 & 78.7/62.1 & 72.8/60.6 & 75.1/58.3 & 79.0/59.8 & 69.2/59.2 & 77.3/61.8 \\
VECO 2.0 & 88.2/78.4 & 74.2/56.9 & 81.0/63.7 & 81.5/63.6 & 83.1/64.6 & 76.8/60.3 & 80.1/64.4 & 76.0/66.4 & 76.6/60.8 & 80.5/60.8 & 70.1/60.9 & 78.9/63.7 \\ \bottomrule
\end{tabular}
}
\caption{XQuAD F1/EM scores for each language in cross-lingual setting.}
\end{table*}

\begin{table*}[]
\centering
\resizebox{1\textwidth}{!}{
\begin{tabular}{@{}lcccccccc@{}}
\toprule
Model & en & ar & de & es & hi & vi & zh & Avg. \\ \midrule
XLM-R & 83.5/70.6 & 66.6/47.1 & 70.1/54.9 & 74.1/56.6 & 70.6/53.1 & 74.0/52.9 & 62.1/37.0 & 71.6/53.2 \\
VECO & 83.6/70.5 & 65.0/44.6 & 69.8/54.6 & 73.8/55.6 & 69.1/51.4 & 73.1/51.8 & 67.3/43.6 & 71.7/53.2 \\
VECO 2.0 & 84.1/71.4 & 74.3/56.3 & 70.3/54.9 & 66.5/46.5 & 71.5/53.7 & 74.2/53.1 & 67.9/43.7 & 72.7/54.2 \\ \bottomrule
\end{tabular}
}
\caption{MLQA F1/EM scores for each language in cross-lingual setting.}
\end{table*}

\begin{table*}[]
\centering
\resizebox{1\textwidth}{!}{
\begin{tabular}{@{}lcccccccccc@{}}
\toprule
Model & en & ar & bn & fi & id & ko & ru & sw & te & Avg. \\ \midrule
XLM-R & 71.5/56.8 & 67.6/40.4 & 64.0/47.8 & 70.5/53.2 & 77.4/61.9 & 31.9/10.9 & 67.0/42.1 & 66.1/48.1 & 70.1/43.6 & 65.1/45.0 \\
VECO & 71.3/58.2 & 73.1/52.8 & 58.9/42.5 & 70.9/55.1 & 77.2/60.0 & 54.2/39.9 & 66.1/37.6 & 65.8/45.7 & 70.6/50.7 & 67.6/49.1 \\
VECO 2.0 & 73.0/60.7 & 73.7/55.8 & 63.6/46.9 & 72.8/56.9 & 79.6/66.7 & 60.5/47.8 & 67.7/39.9 & 73.2/60.1 & 75.6/57.1 & 71.1/54.7 \\ \bottomrule
\end{tabular}
}
\caption{TyDiQA F1/EM scores for each language in cross-lingual setting.}
\end{table*}

\begin{table}[]
\centering
\small
\begin{tabular}{@{}lccccc@{}}
\toprule
Model & de & fr & ru & zh & Avg. \\ \midrule
XLM-R & 67.5 & 66.5 & 73.5 & 56.7 & 66.0 \\
VECO & 89.6 & 84.6 & 87.4 & 78.5 & 85.0 \\
VECO 2.0 & 90.5 & 86.1 & 88.1 & 80.2 & 86.2 \\ \bottomrule
\end{tabular}
\caption{BUCC F1 scores for each language in cross-lingual setting.}
\end{table}

\begin{table*}[]
\centering
\resizebox{1\textwidth}{!}{
\begin{tabular}{@{}lccccccccccccccccccc@{}}
\toprule
Model & af & ar & bg & bn & de & el & es & et & eu & fa & fi & fr & he & hi & hu & id & it & ja &  \\ \midrule
XLM-R & 58.2 & 47.5 & 71.6 & 43.0 & 88.8 & 61.8 & 75.7 & 52.2 & 35.8 & 70.5 & 71.6 & 73.7 & 66.4 & 72.2 & 65.4 & 77.0 & 68.3 & 60.6 &  \\
VECO & 48.2 & 70.9 & 86.7 & 57.7 & 97.5 & 81.5 & 94.8 & 89.7 & 62.9 & 82.1 & 87.9 & 88.8 & 74.7 & 80.7 & 87.6 & 89.6 & 89.2 & 83.2 &  \\
VECO 2.0 & 84.0 & 73.4 & 87.4 & 68.2 & 98.1 & 82.8 & 94.8 & 80.9 & 58.3 & 91.6 & 91.9 & 91.4 & 80.2 & 92.0 & 89.3 & 92.5 & 87.8 & 89.0 &  \\
\midrule
 & jv & ka & kk & ko & ml & mr & nl & pt & ru & sw & ta & te & th & tl & tr & ur & vi & zh & Avg. \\
 \midrule
XLM-R & 14.1 & 52.1 & 48.5 & 61.4 & 65.4 & 56.8 & 80.8 & 82.2 & 74.1 & 20.3 & 26.4 & 35.9 & 29.4 & 36.7 & 65.7 & 24.3 & 74.7 & 68.3 & 57.3 \\
VECO & 17.6 & 58.5 & 53.9 & 75.3 & 80.1 & 64.2 & 94.4 & 92.8 & 88.6 & 37.4 & 61.9 & 65.8 & 84.5 & 52.5 & 89.3 & 64.3 & 85.8 & 82.7 & 75.1 \\
VECO 2.0 & 36.1 & 76.3 & 69.6 & 86.3 & 87.5 & 78.9 & 94.5 & 93.2 & 92.2 & 35.1 & 68.7 & 80.8 & 88.3 & 64.7 & 92.9 & 78.5 & 94.7 & 93.3 & 81.8 \\ \bottomrule
\end{tabular}
}
\caption{Tatoeba accuracy scores for each language in cross-lingual setting.}
\end{table*}

\end{document}